\documentclass[conference]{IEEEtran}
\IEEEoverridecommandlockouts

\usepackage{amsmath,amssymb,amsfonts}
\usepackage{algorithmic}
\usepackage{graphicx}
\usepackage{textcomp}
\usepackage{xcolor}

\usepackage{xcolor,colortbl}
\usepackage[utf8]{inputenc} 
\usepackage{hyperref}       
\usepackage{url}            
\usepackage{booktabs}       
\usepackage{amsfonts}       
\usepackage{nicefrac}       
\usepackage{microtype}      
\usepackage{amsmath}
\usepackage{algorithm}
\usepackage{algorithmic}
\usepackage{float}
\usepackage{multirow, adjustbox}
\usepackage{booktabs, makecell, arydshln, rotating}
\usepackage{pifont}
\usepackage{stfloats}
\usepackage{arydshln}
\usepackage{tcolorbox}
\tcbuselibrary{breakable}
\renewcommand{\algorithmiccomment}[1]{\bgroup\hfill\tiny//~#1\egroup}
\newcommand{\cmark}{\textcolor{cyan}{\ding{51}}}%
\newcommand{\xmark}{\textcolor{orange}{\ding{55}}}%

\def\BibTeX{{\rm B\kern-.05em{\sc i\kern-.025em b}\kern-.08em
    T\kern-.1667em\lower.7ex\hbox{E}\kern-.125emX}}
\begin{document}

\title{FedHFT: Efficient Federated Fine-tuning with Heterogeneous Edge Clients}

\author{\IEEEauthorblockN{
Fatih Ilhan\textsuperscript{1}, 
Selim Furkan Tekin\textsuperscript{1}, 
Tiansheng Huang\textsuperscript{1}, 
Gaowen Liu\textsuperscript{2}, 
Ramana Kompella\textsuperscript{2}, \\
Greg Eisenhauer\textsuperscript{1}, 
Yingyan Celine Lin\textsuperscript{1}, 
Calton Pu\textsuperscript{1}, 
Ling Liu\textsuperscript{1}}
\textsuperscript{1}\textit{School of Computer Science, Georgia Institute of Technology, USA}, \\
\textsuperscript{2}\textit{CISCO Research, USA}
}

\maketitle

\begin{abstract}
  Finetuning pre-trained large language models (LLMs) has become a common practice for personalized natural language understanding (NLU) applications on downstream tasks and domain-specific datasets. However, there are two main challenges: (i) limited and/or heterogeneous data for fine-tuning due to proprietary data confidentiality or privacy requirements, and (ii) varying computation resources available across participating clients such as edge devices. This paper presents FedHFT - an efficient and personalized federated fine-tuning framework to address both challenges. First, we introduce a mixture of masked adapters to handle resource heterogeneity across participating clients, enabling high-performance collaborative finetuning of pre-trained language model(s) across multiple clients in a distributed setting, while keeping proprietary data local. Second, we introduce a bi-level optimization approach to handle non-iid data distribution based on masked personalization and client clustering. Extensive experiments demonstrate significant performance and efficiency improvements over various natural language understanding tasks under data and resource heterogeneity compared to representative heterogeneous federated learning methods.
\end{abstract}


\section{Introduction}
The trend of training increasingly larger transformer-based language models over big datasets has brought exceptional performance across many language tasks~\cite{t5,gpt3,llama2}. These \textit{foundation} models can be further optimized for the downstream tasks defined by users through finetuning on domain-specific datasets~\cite{bert}. Although finetuning improves the applicability and performance, it requires high computation capability, network availability and storage of an adequate amount of data, which may not be suitable for all users due to limited resources~\cite{adapter,recap}. To democratize the usage of LLMs, it is crucial to address the challenges of privacy-aware finetuning under resource constraints by exploring personalized and efficient collaborative finetuning frameworks and algorithms. 

To this end, we propose an efficient federated finetuning framework with a mixture of masked adapters, coined as FedHFT. Federated learning enables a distributed population of participants (e.g. edge devices) to jointly train a global model while allowing each participant to keep their proprietary data local and utilize distributed resources by sharing the model instead of centralizing data and computation~\cite{fedavg}. 
We augment this framework for finetuning, aiming to support privacy-aware and personalized finetuning methodology to encourage collaboration across a population of disparate clients with heterogeneous resources (CPU, GPU, memory, network, etc.) and non-iid data distribution.
We propose utilizing a mixture of masked adapters in federated setting to improve memory efficiency during finetuning, reduce communication costs, and handle data heterogeneity.

Adapters as a parameter-efficient finetuning technique enable reducing the memory footprint of the finetuning process by only updating a set of injected additional weights while keeping the backbone frozen~\cite{adapter}. In particular, LoRA has gained attention due to the efficient representation of model updates in low-rank~\cite{lora}, which makes it attractive for the federated setting in our work due to the lower memory footprint of finetuning in clients and lower communication cost. However, the challenges related to data heterogeneity are still relevant while using adapters. 

For this propose, we utilize a bi-level approach based on personalization with client clustering and masked updates to handle data heterogeneity across participating clients. First, we define a mixture of adapters instead of aggregating the updates over one global set of adapter weights. After receiving the client updates, we fit a Gaussian Mixture Model over updates after dimension reduction using PCA to cluster clients. Each client contributes to the corresponding adapter at each cluster based on its likelihood of belonging to the cluster. To further enhance personalization and reduce the communication cost, clients approximate how much each dimension in the adapter weights contributes to the model output using Fisher information and only communicate the dimensions with high importance. We conduct extensive experiments on various language tasks with multiple models and observe up to 3.1x memory and 136.9x communication cost reduction compared to baseline federated learning algorithm~\cite{fedavg} while having superior performance.

\begin{figure*}[t]
    \centering
    \includegraphics[width=0.85\textwidth]{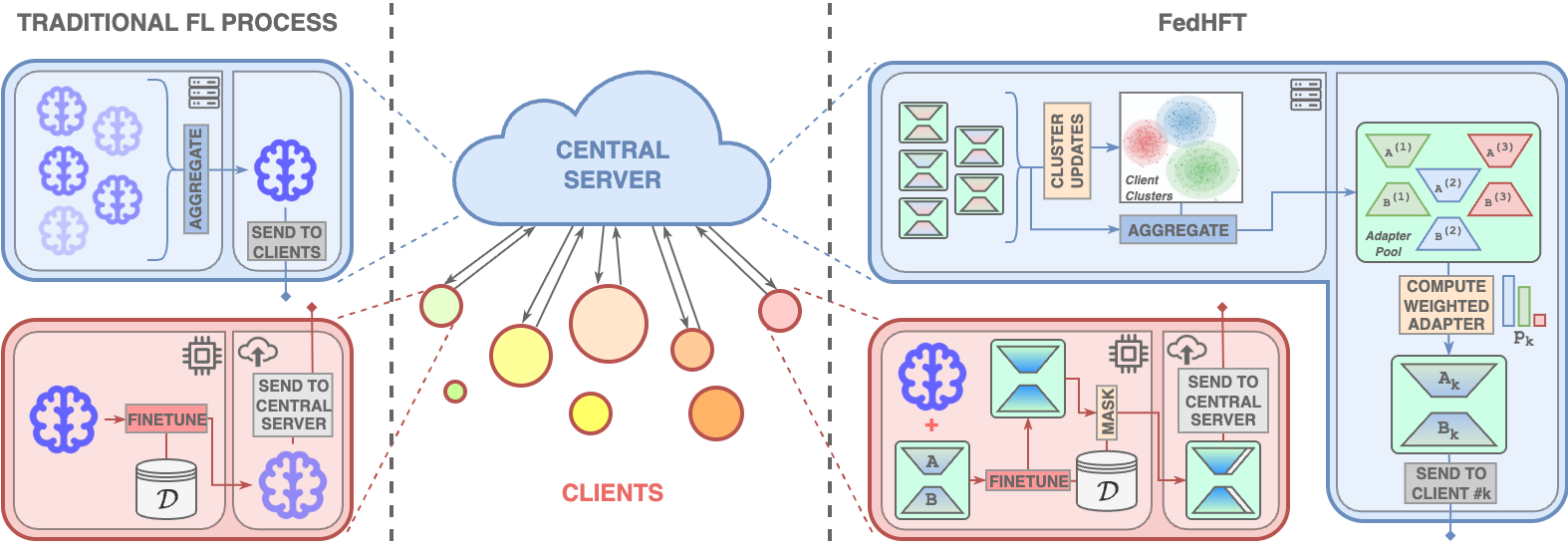}
    \vspace{-0.2cm}
    \caption{System architecture of FedHFT. At each round, the central server computes the adapter weights for each client based on cluster assignment scores. After receiving the adapter weights, participating edge clients initiate local finetuning over their local database and masks based on Fisher information before uploading them back. Lastly, the central server receives the updated adapter weights, updates the adapter weights of each cluster and cluster assignment scores.}
    \label{fig:arch}
\end{figure*}

\section{Related Work}
\label{sec:app_rw}
Recent techniques such as prefix tuning~\cite{prefix, fedprompt} and few-shot inference~\cite{llama2} enable deploying LLM applications without finetuning the model, particularly when the data and hardware are limited. However, finetuning is necessary for mission-critical applications,  where high performance is desired. Parameter-efficient finetuning techniques enable model finetuning with a significantly lower memory by reducing the number of trainable parameters~\cite{peft, recap}. In particular, adapters are widely adopted for LLMs due to high memory efficiency~\cite{adapter, llmadapters}.

Federated learning (FL) enables collaboration among distributed clients while keeping private data local~\cite{fedavg,mofl}. FedAvg as the first FL algorithm averages the weight updates received from clients at each round~\cite{fedavg}. Several works such as FedProx~\cite{fedprox} and SCAFFOLD~\cite{scaffold} propose modified objectives and correction steps to handle non-iid data distribution across clients. Another direction involves grouping clients into clusters. Several works such as FedCAM~\cite{fedcam}, and IFCA~\cite{ifca} group clients based on updates or performance and optimize multiple models to achieve better personalization. However, these approaches fail to handle resource heterogeneity across clients as they assume all participating clients have similar computation, memory, and communication resources, which is usually not the case in practice~\cite{scalefl}. 

To address resource heterogeneity, gradient compression~\cite{dgc} or sparsification~\cite{grad_sparse} techniques reduce the communication cost for clients with slow networks but still demand high memory during finetuning and also require storing the residual gradients. Certain recent techniques~\cite{scalefl, heterofl, scalefl_2} focus on model partitioning to reduce finetuning and communication costs but do not consider data heterogeneity. 
In our work, instead of optimizing the entire model, we utilize adapters by injecting a set of trainable parameters while keeping the backbone frozen. The smaller size of weight updates reduces the memory and communication cost, which motivates our choice of utilizing LoRA~\cite{lora}. 
Some recent FL techniques also utilize LoRA for resource efficiency such as in \cite{hetlora}, different ranks are used for each client, where ranks are sampled from a uniform or power-law distribution, and in \cite{fl_lora} only the zero-initialized component of LoRA is updated to reduce communication costs and improve stability. However, neither of these approaches takes the data heterogeneity into account.

\section{Methodology}
In this section, we first provide background on federated finetuning and introduce the challenges due to data and resource heterogeneity. Then, we present our approach that enables federated finetuning of large language models by employing a mixture of masked adapters.

\subsection{Federated Fine-tuning}
Consider a federated learning system of $K$ clients, each has its private local dataset of size $N_k$, denoted by $\mathcal{D}_k = \{(x_i, y_i)\}_{i=1}^{N_k}$, where $(x_i, y_i)$ is the $i$th data sample and label, for each client $k \in \{1, \hdots, K\}$. Instead of training a global model from scratch as in traditional FL, the goal is finetuning an already pre-trained model with weights $\theta_0$ over $T$ rounds to minimize the following:
\noindent
\begin{equation}
    \min_\theta f(\theta) = \sum_{k=1}^N q_k F_k(\theta),
\end{equation}
\noindent
where $F_k := \mathbb{E}_{(x, y) \sim \mathcal{D}_k}[\ell_k(\theta; x, y)]$ with $\ell_k$ as the loss of client-$k$ and $q_k=\frac{N_k}{\sum_{k'=1}^N N_{k'}}$.

With FedAvg as a baseline for fine-tuning, at each round $t$, the global model weights are distributed to participating clients $\mathcal{S}_t$. Each client $k \in \mathcal{S}_t$ performs $E$ epochs of local fine-tuning, wherein an optimization step with learning rate $\eta$ is performed for each batch $b \in \mathcal{D}_k$: $\theta \leftarrow \theta - \eta \nabla \ell(\theta; b)$. After completion of local finetuning, each client $k$ returns the updated weights $\theta_{t+1}^k$ (or updates $\Delta \theta_{t+1}^k := \theta_{t+1}^k - \theta_t$) to the central server. The server then aggregates the received updates and obtains the updated global model weights to start the next round of federated learning: 
\begin{equation}
    \theta_{t+1} \leftarrow \theta_t + \sum_{k \in \mathcal{S}_t} \frac{N_k}{\sum_{k' \in \mathcal{S}_t} N_{k'}} \Delta \theta_{t+1}^k.
\end{equation}

\noindent 
\subsubsection{Resource Heterogeneity}

Each client may have varying capabilities w.r.t. computation power, RAM, GPU memory, network bandwidth etc. This situation can cause challenges for certain clients with limited resources while finetuning the full model weights and communicating them between the central server, which can require hundreds of GBs of GPU memory and downloading/uploading billions of parameters at each communication round depending on the model size. 

Gradient compression techniques can achieve lower communication costs by reducing the size of the uploaded model weight updates to the central server through gradient masking and encoding. For instance, Deep Gradient Compression (DGC)~\cite{dgc} achieves this by communicating the gradients only once they reach a certain magnitude threshold over FL rounds, i.e., $\Delta \theta_{t+1}^k \leftarrow \Delta \theta_{t+1}^k \odot (|\theta_{t+1}^k| > \tau)$. This technique can reduce the communication cost per round, but clients still need to fine-tune the full model weights.

Adapter-based parameter-efficient finetuning techniques can achieve lower memory footprint and also lower communication cost by reducing the number of trainable parameters and hence, the size of the updates to communicate. We utilize LoRA~\cite{lora} in the federated setting by approximating the weight matrix updates such that $\Delta W_{t+1}^k \sim B_k A_k$, where $B_k$ and $A_k$ are low-rank components but, we still need to address the challenges related to data heterogeneity.

\subsubsection{Data Heterogeneity}

When learning over with data heterogeneity across clients, i.e. the scenario when the local dataset $\mathcal{D}_k$ at each client $k$ may have different feature/target distributions or dataset sizes,  adapting to varying client behaviors is desired. Notably, FedProx introduces a proximal term to local optimization objectives such that each local solver minimizes the following objective: $F_k(\theta) + \frac{\mu}{2}||\theta-\theta_t||^2$ to prevent client drift while also allowing clients to perform variable amounts of computation. SCAFFOLD targets the same problem by utilizing control variates to correct local updates: $\theta \leftarrow \theta - \eta(\nabla \ell(\theta; b) + \Delta c)$, where $\Delta c$ is the client correction term, whose computation requires additional pass over the model and communication between the server.

Lastly, multi-center FL also aims to tackle the challenge of data heterogeneity by clustering the participating clients and optimizing one model for each cluster instead of one global model, resulting in the following objective: 
\begin{equation}
    \min_{\{\theta^{(c)}\}_{c=1}^C} f(\{\theta^{(c)}\}_{c=1}^C) = \sum_{k=1}^N q_k \sum_{c=1}^C p_{kc}  F_k(\theta^{(c)}), \label{eq:cam}
\end{equation}
where,  $q_k=\frac{N_k}{\sum_{k'=1}^N N_{k'}}$ and $p_{kc}=1$ if the $k$th client is assigned to the $c$th cluster and zero otherwise. We consider update similarity as the clustering criteria and utilize Gaussian Mixture Modeling (GMM) to perform clustering.

\subsection{Mixture of Masked Adapters}

We address the data and resource heterogeneity problem in federated learning with a mixture of masked adapters. First, we achieve a low memory footprint during local finetuning and reduce the communication cost by utilizing the technique of low-rank approximation for weight updates. To handle data heterogeneity, we group clients into clusters based on weight updates and learn a mixture of adapters. The weight updates received from each client contribute to the aggregation step in the central server based on its cluster membership scores. Lastly, we allow clients to mask their updates to enhance communication efficiency and personalization. In this way, only the subnetworks with the most effect on the objective loss on their local datasets will receive updates. We illustrate the workflow in Figure~\ref{fig:arch}. 

\subsubsection{Initialization} 
At the beginning of the process, we initialize the adapter weights for each cluster. Specifically, we follow the low-rank approximation-based adapter technique proposed in LoRA~\cite{lora}. For notational simplicity, we do not use any subscript to indicate the weight matrix that the adapter weights correspond to. We can approximate any weight update matrix $\Delta W \in \mathbb {R}^{d_{out}\times d_{in}}$ as $\Delta W \sim BA$, where $B \in \mathbb{R}^{d_{out} \times r}$ and $A \in \mathbb{R}^{r \times d_{in}}$ are the low-rank components with rank $r$. We set $B$ to a zero matrix and sample the elements of $A \sim \mathcal{N}(0, \sigma^2)$ (Algorithm~\ref{alg_main}, line 1). For the sake of simplicity, let  $U = \{U_c\}_{c=1}^C = \{(B_c, A_c)\}_{c=1}^C$ denote the set of adapter weights for any weight matrix in attention blocks, where $C$ is the number of clusters.

\begin{algorithm}[t!]
\algsetup{linenosize=\normalsize}
\textbf{Parameters}: number of communication rounds $T$, number of warmup rounds $T_w$, number of local training epochs $E$, number of clusters $C$, client set $\mathcal{S}$ with $K$ clients \\
\textbf{Inputs}: Dataset $\mathcal{D}_k = \{(x_i, y_i)\}_{i=1}^{N_k}$ for each client $k$, pre-trained model with weights $\theta$ \\
\textbf{Outputs}: The set of adapter weights $\mathcal{U} = \{(B_c, A_c)\}_{c=1}^C$ and cluster assignment scores $P$
\caption{\texttt{FedHFT}}
	\begin{algorithmic}[1]
	\STATE \textbf{Initialize} adapter weights for each cluster $c$:  $B_c^{(0)}=0, A_c^{(0)} \sim \mathcal{N}(0, \sigma^2)$
        \STATE \textbf{Initialize} cluster assignment scores $P=[p_{kc}] \in \mathbb{R}^{K \times C}$ such that $p_{kc} = \frac{1}{C} \; \forall \; c, k$
	\FOR{round $t = 0, \hdots T-1$}
	    \FOR{client $k \in \mathcal{S}$ in parallel}
                \STATE \textbf{Merge} adapters into model with Eq.~\eqref{eq:init}
	        \STATE \textbf{Finetune} $\theta_k$ for $E$ epochs and obtain adapter weights $\Tilde{B}_k, \Tilde{A}_k$ s.t. $\Delta \theta_k         \approx \Tilde{B}_k\Tilde{A}_k$
                \STATE \textbf{Mask} updates based on empirical Fisher with Eq.~\eqref{eq:imp_grad} and obtain masked low-rank weight updates $\bar{B}_k, \bar{A}_k$
	    \ENDFOR
            \FOR{cluster $c = 1, \hdots C$}
	        \STATE \textbf{Aggregate} updates and obtain new adapter weights with Eq.~\eqref{eq:aggr}
	    \ENDFOR
            \IF{$t \geq T_w$}
            \STATE \textbf{Update} cluster assignment scores $P$ with Eq.~\eqref{eq:update_gmm}
            \ENDIF
	\ENDFOR
	\RETURN $\mathcal{U}$, $P$
	\end{algorithmic} \label{alg_main}
 
\end{algorithm}

We define cluster assignment scores such that $p_{kc}$ represents the probability that the $k$th client belongs to the $c$th cluster instead of hard binary assignment in Eq.~\ref{eq:cam}. We set these values to $\frac{1}{C}$ for each client at initialization as we assume no prior information about clients (Algorithm~\ref{alg_main}, line 2). We keep these values constant through the first $T_w$ warmup round (i.e. no clustering is performed) to eliminate instability in early rounds.

Then, at each FL round $t$, instead of sending the full global model weights to clients, we communicate a weighted mixture of weights considering the cluster assignment score of clients. The initialization of the local model weights $\theta_k$ for client-$k$ is performed by merging the mixture of adapter weights of each cluster-$c$ (Algorithm~\ref{alg_main}, line 5):

\noindent
\begin{equation}
    \theta_k \leftarrow \theta + \sum_{c=1}^C p_{kc} B_c^{(t)}A_c^{(t)},
    \label{eq:init}
\end{equation}
\noindent
i.e., each client starts the local optimization from a personalized point based on its cluster assignment scores.

\subsubsection{Local Finetuning and Update Masking}
During local finetuning, we only update the adapter and classifier weights while keeping the rest of the model frozen to reduce memory cost (Algorithm~\ref{alg_main}, line 6). To further reduce the communication cost and enhance personalization, we perform masking over the adapter weights (Algorithm~\ref{alg_main}, line 7). To this end, we approximate how much the model output would change if we change the weight $\theta_k$ to $\theta_k+\delta$, where $\delta$ is a small weight perturbation by estimating the KL divergence between two output distributions~\cite{sparse_train}:
\noindent
\begin{equation}
    \mathbb{E}_{x}D_{KL}(p_{\theta_k}(y|x) || p_{\theta_k+\delta}(y|x)) \approx \delta^2 F_{\theta_k} + O(\delta^3), \nonumber
\label{eq:fisher_1}
\end{equation}
\noindent
where $F_{\theta_k}$ is the $k$th diagonal value of the Fisher information matrix and is defined as:
\begin{equation}
    \scalebox{1}{$F_{\theta_k} = \mathbb{E}_{x\sim p(x)}\Big[\mathbb{E}_{y\sim p(y|x)} \Big(\frac{\partial {\log p_{\theta_k}(y|x)}}{\partial\theta_k}\Big)^2\Big]$}.
\label{eq:fisher_2}
\end{equation}
\noindent
In other words, lower this metric is for any weight, it is less crucial to update it as its impact on the model output distribution is low. Then, we compute the importance of each output dimension $d$ for each attention weight $W_k$ based on the total approximated expectation in Eq.\eqref{eq:fisher_2} (also referred to as empirical Fisher~\cite{sparse_train}):
\begin{align}
    \mathcal{I}_k[d] &= \sum_{d' \in \{1, ... d_{in}\}} \Big(\frac{\partial {\log p_{\theta_k}(\mathcal{D}_s)}}{\partial W_k [d, d']}\Big)^2, \\
    & \propto \sum_{d' \in \{1, ... d_{in}\}} \langle B_k[d, :], A_k[:, d'] \rangle^2,
    \label{eq:imp_grad}
\end{align}
where the inner product of the $d$th row of $B_k$ and the columns of $A_k$ are used to approximate the derivative of log-likelihood to the weights corresponding to $d$th hidden dimension. Then, we mask the dimensions within bottom-$r\%$ in terms of the importance value in Eq.~$\eqref{eq:imp_grad}$ (i.e. the dimensions with a low average of gradient magnitudes). We denote the masked adapter weights for client-$k$ as $\bar{B}_k, \bar{A}_k$. In practice, we only communicate the unmasked dimensions with the central server, further reducing communication costs.

\begin{figure}[t!]
    \centering
    \includegraphics[width=0.75\columnwidth]{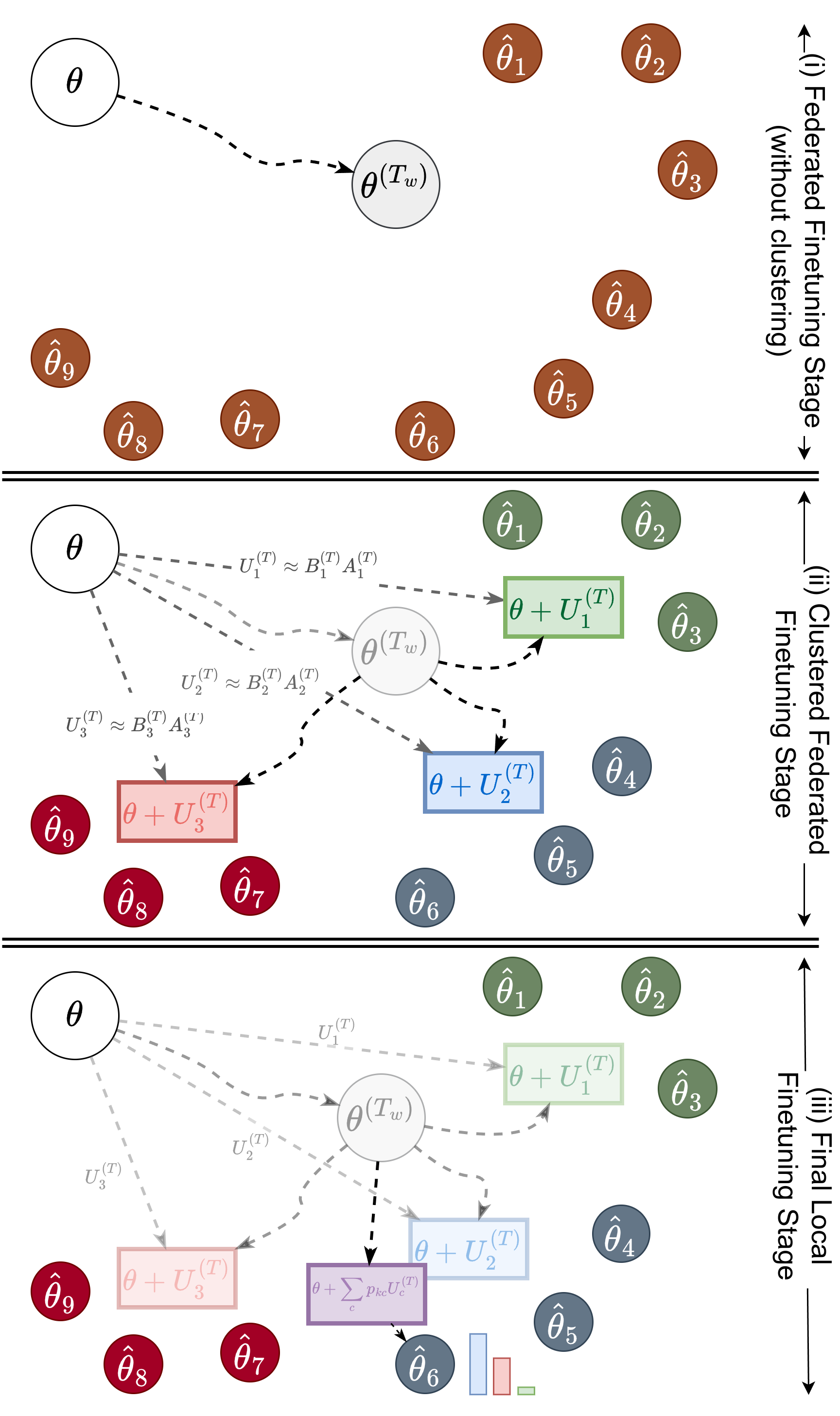}
    \caption{Three main stages in our framework: (i) we perform federated finetuning without clustering for $T_w$ warmup rounds, (ii) we continue the process with a mixture of adapters, where each adapter tracks the updates of the corresponding cluster, (iii) each client can perform a final finetuning on its local data using the initialization point ($\theta+\sum_{c} p_{kc} B_c^{(T)}A_C^{(T)}$) before deployment for inference.}
    \label{fig:stages}
\end{figure}

\subsubsection{Aggregation and Client Clustering}
During the aggregation of adapter weights at round $t$, we compute a weighted average of the received updates at each cluster $c$ as below:
\begin{equation}
    B_c^{(t+1)}, A_c^{(t+1)} \leftarrow \texttt{SVD}(B_c^{(t)}A_c^{(t)} + \sum_{k} p_{kc} \bar{B}_k \bar{A}_k),
    \label{eq:aggr}
\end{equation}
where we allow each client-$k$ to contribute to the updates for cluster-$c$ based on $p_{kc}$ (Algorithm~\ref{alg_main}, line 10). After performing the aggregation over full-rank updates, we convert it to low-rank representation using singular-value decomposition (SVD), and scale each column of the first low-rank component with the computed singular values. For the final dense layer weights, the averaging procedure using $p_{kc}$ as the cluster assignment is the same except, there is no need to reconstruct updates from low-rank updates and perform SVD operation as we do not apply LORA for these weight matrices.

Before starting the next round $t$, if $t \geq T_w$, i.e. warmup stage is completed, we update the cluster assignments based on the received weight updates using Gaussian Mixture Modeling (GMM):
\begin{equation}
    p_{kc} = \frac{\alpha_{c} \phi(\boldsymbol{u}_k | \boldsymbol{\mu}_c, \boldsymbol{\Sigma}_c)}{\sum_{c'=1}^C \alpha_{c'}\phi(\boldsymbol{u}_k | \boldsymbol{\mu}_{c'}, \boldsymbol{\Sigma}_{c'})},
    \label{eq:update_gmm}
\end{equation}
where $\boldsymbol{u}_k$ is the vectorized weight updates of the $k$th client used for clustering, and $\phi(\boldsymbol{u} | \boldsymbol{\mu}, \boldsymbol{\Sigma}) = \frac{1}{(2\pi)^{d/2} \sqrt{|\boldsymbol{\Sigma}|}}\exp{(-\frac{1}{2}(\boldsymbol{u}-\boldsymbol{\mu})^T \boldsymbol{\Sigma}^{-1}(\boldsymbol{u}-\boldsymbol{\mu}))}$. We follow the expectation-maximization (EM) algorithm to estimate the GMM parameters. After completing $T$ rounds of federated finetuning, each client can also perform a final finetuning on its local data using the initialization point ($\theta+\sum_{c} p_{kc} B_c^{(T)}A_c^{(T)}$) before deployment for inference. We also provide Figure~\ref{fig:stages} to illustrate the entire optimization process.

\section{Results}

In this section, we present the results obtained from extensive experiments performed on nine datasets. We show that FedHFT consistently performs better or competitively compared to recent representative techniques with significantly lower memory and communication costs. Main results are reported in Tables~\ref{tab:main}, \ref{tab:t5_base} and \ref{tab:t5_large}. We also analyze the behavior of our approach under various data and resource heterogeneity settings and report the ablation results. We describe the datasets, preprocessing, model architectures, compared methods, implementation details, and the federated finetuning setup in Section~\ref{sec:setup}.

\subsection{Experiment Setup}
\label{sec:setup}

\subsubsection{Datasets}

We consider datasets with varying sizes (2.5k to 400k samples), domains (news, fiction, art, Wikipedia), and tasks (question answering, sentiment analysis, question answering etc.). We experiment on the following six datasets from the GLUE benchmark~\cite{glue}: The Corpus of Linguistic Acceptability (CoLA)~\cite{cola} with 10.7k/1.04k/10.6k, The Stanford Sentiment Treebank (SST-2)~\cite{sst2} with 67.3k/872/1.82k, The Microsoft Research Paraphrase Corpus (MRPC)~\cite{mrpc} with 3.67k/408/1.73k, The Quora Question Pairs (QQP)~\cite{glue} with 364k/40.4k/391k, The Multi-Genre Natural Language Inference Corpus (MNLI)~\cite{mnli} with 393k/9.8k/9.8k, The Recognizing Textual Entailment (RTE)~\cite{glue} with 2.49k/277/3k, and also the AG's news topic classification dataset (AgNews)~\cite{agnews} with 120k/-/7.6k train, validation and test samples. We also evaluate on SQuADv1~\cite{squad} containing 100000+ question-answering samples from 500+ texts (80\%, 10\%, 10\%) for train, validation and test. Due to the high number of experiment runs and the public unavailability of test labels, we keep 10\% of the train samples for hyperparameter tuning and report the results on the validation sets. For tokenization, we use the pre-trained tokenizers provided by the open-source HuggingFace library.

\subsubsection{Implementation Details}
We collect results with the two transformer-based LLM architectures: (i) encoder-only BERT~\cite{bert} model, particularly the base and large versions, respectively having 12/24 layers, 768/1024 hidden size, 12/16 heads, and in total 110M/340M parameters, (ii) encoder-decoder T5~\cite{t5} model with the base and large versions, respectively having 12/24 encoder layers, 12/24 decoder layers, 768/1024 hidden size, 12/16 heads, and in total 220M/770M parameters. 

\begin{figure}[t!]
    \centering
    \includegraphics[width=\columnwidth]{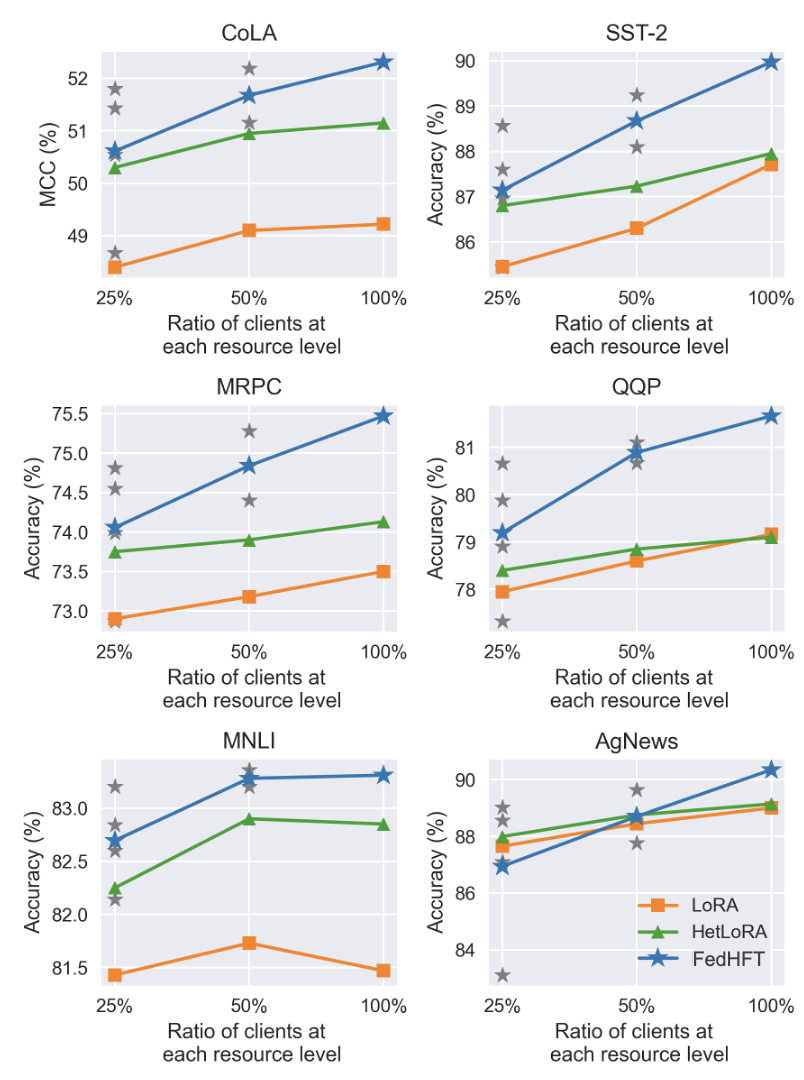}
    \caption{Performance of FedHFT on various resource heterogeneity settings with BERT-base, $K=50$, $\alpha=5$ and the ratio of clients with adapter rank $r \in [4, 8, 16, 32]$. We plot the average performance of each client group corresponding to the same adapter rank with FedHFT (grey scattered stars) and the average performance of all clients for FedHFT (blue) and other resource-efficient techniques LoRA (yellow) and HetLoRA (green).}
    \label{fig:res_het}
\end{figure}

For comparisons, we consider FedAvg~\cite{fedavg} as our baseline where we aggregate the weight updates received from clients at each round by averaging the values based on data quantity. As the methods that address data heterogeneity, we compare our approach with FedProx~\cite{fedprox}, SCAFFOLD~\cite{scaffold} and FedCAM~\cite{fedcam}, where we finetune all model parameters for fair comparison as these methods can only rely on full model training/finetuning. We also perform comparisons with DGC~\cite{dgc} (50\% compression), finetuning with LoRA~\cite{lora} ($r=32$) and HetLoRA~\cite{hetlora} ($r_{min}=5, r_{min}=50, \gamma = 0.99$, and tuned accordingly for different resource level comparisons in Figure~\ref{fig:res_het}) for resource heterogeneity. As methods that rely on model partitioning to handle resource heterogeneity, we consider HeteroFL~\cite{heterofl} and ScaleFL~\cite{scalefl} with four resource levels, respectively allowing operation on the 25\%, 50\%, 75\% and 100\% of the full model.

For the federated finetuning setup, we consider an environment with $K=50$ participating clients. We randomly sample data for each client using a Dirichlet distribution over labels with concentration parameter $\alpha=5$ to simulate data heterogeneity for classification tasks. For question answering, we apply heterogeneity by assigning texts to clients (i.e., QA pairs of a text are not distributed across multiple clients). We set $T=20$, $E=2$, and evaluate with the checkpoint that has the best overall validation performance. For experiments with $r_a<1$, we scale the number of rounds by $\frac {1}{r_a}$. For our approach, we set the number of warmup rounds $T_w=5$ and apply adapters for query and value weight matrices, and set the rank to $r=32$. We perform clustering over the classifier weight updates for classification with BERT and final dense layer weight updates for text generation tasks with T5 (Updates are considered relative to the mean of weights across clusters at the beginning of the round). We set the number of clusters to $C=3$ as fixed but, our approach is robust to this parameter for up to ten clusters based on our experiments. The number of clusters can also be determined and even dynamically adjusted during the process using various information criteria (e.g. AIC, BIC). We set the masking ratio $r=0.5$. We set the batch size to $32$ and perform a search for learning rate $[1e-5, 1e-2]$ in all experiment runs.
Our experiments are conducted with A100-40GB GPU and 16GB RAM as the central server in federated finetuning simulations. We measure the latency of edge finetuning with RTX3060-12GB GPU and assume a 100 Mbps connection between clients and the central server. We use Python 3.9 with PyTorch 2.2.

\begin{table*}[]
\centering
\begin{adjustbox}{width=0.85\textwidth}
\begin{tabular}{clccccccccc}
\toprule
\multirow{2}{*}{Model} & \multirow{2}{*}{\begin{tabular}[c]{@{}l@{}}Finetuning \\ Method\end{tabular}} & \multicolumn{2}{c}{Cost (GB) ($\downarrow$)} & \multicolumn{7}{c}{Performance on Dataset ($\uparrow$)} \\
 &  & Memory & Comm. & CoLA & SST-2 & MRPC & QQP & MNLI  & RTE & AgNews \\ \midrule
\multirow{7}{*}{BERT-base} & \textbf{FedAvg~\cite{fedavg}} & $1.98$ & $17.76$ & $49.55$ & $88.16$ & $73.75$ & $78.61$ & $81.60$ & $56.23$ & $89.10$ \\
 & \textbf{FedProx~\cite{fedprox}} & $1.98$ & $17.76$ & $\underline{52.14}$ & $89.36$ & $\boldsymbol{75.73}$ & $80.86$ & $82.31$ & $60.28$ & ${90.11}$ \\
 & \textbf{SCAFFOLD~\cite{scaffold}} & $1.98$ & $35.52$ & $50.11$ & $\underline{89.50}$ & $75.00$ & $80.32$ & $82.88$ & $61.45$ & $90.01$ \\
 & \textbf{FedCAM~\cite{fedcam}} & $1.98$ & $17.76$ & $50.75$ & $88.51$ & $75.03$ & $\underline{80.94}$ & $\underline{83.10}$ & $\underline{61.74}$ & $89.74$ \\
 \cdashline{2-11}\noalign{\vskip 0.4ex}
 & \textbf{DGC~\cite{dgc}} & $1.98$ & $8.96$ & $49.00$ & $86.47$ & $73.99$ & $78.53$ & $81.03$ & $56.13$ & $90.00$ \\
 & \textbf{HeteroFL~\cite{heterofl}} & $1.24$ & $11.10$ & $49.04$ & $87.22$ & $72.10$ & $78.00$ & $82.05$ & $57.77$ & $\underline{90.90}$ \\
 & \textbf{ScaleFL~\cite{scalefl}} & $1.22$ & $10.95$ & $49.24$ & $87.18$ & $72.75$ & $78.17$ & $81.66$ & $58.90$ & $\boldsymbol{91.24}$ \\
 & \textbf{LoRA~\cite{lora}} & $\boldsymbol{0.64}$ & $\underline{0.20}$ & $49.22$ & $87.71$ & $73.50$ & $79.16$ & $81.47$ & $56.40$ & $89.00$ \\
  & \textbf{HetLoRA~\cite{hetlora}} & $0.68$ & $0.23$ & $51.15$ & $87.95$ & $74.13$ & $79.10$ & $82.85$ & $59.49$ & $89.14$ \\
 \rowcolor{gray!30}
 & \textbf{FedHFT} & $\boldsymbol{0.64}$ & $\boldsymbol{0.14}$ & $\boldsymbol{52.31}$ & $\boldsymbol{89.97}$ & $\underline{75.47}$ & $\boldsymbol{81.66}$ & $\boldsymbol{83.31}$ & $\boldsymbol{62.74}$ & ${90.34}$ \\ \midrule
\multirow{7}{*}{BERT-large} & \textbf{FedAvg~\cite{fedavg}} & $6.21$ & $55.70$ & $51.19$ & $90.62$ & $75.00$ & $83.44$ & $84.11$ & $58.16$ & $91.08$ \\
 & \textbf{FedProx~\cite{fedprox}} & $6.21$ & $55.70$ & $53.69$ & $92.25$ & $\boldsymbol{78.10}$ & $84.88$ & $86.09$ & $\underline{62.10}$ & $\underline{91.85}$ \\
 & \textbf{SCAFFOLD~\cite{scaffold}} & $6.21$ & $111.40$ & $\boldsymbol{53.80}$ & $\boldsymbol{92.66}$ & $77.49$ & $\underline{84.90}$ & $\underline{86.25}$ & $62.10$ & $90.73$ \\
 & \textbf{FedCAM~\cite{fedcam}} & $6.21$ & $55.70$ & $53.66$ & $92.08$ & $77.21$ & $84.67$ & $85.60$ & $61.96$ & $91.78$ \\
 \cdashline{2-11}\noalign{\vskip 0.4ex}
  & \textbf{DGC~\cite{dgc}} & $6.21$ & $28.15$ & $50.02$ & $91.00$ & $75.25$ & $83.01$ & $84.55$ & $58.77$ & $91.07$ \\
  & \textbf{HeteroFL~\cite{heterofl}} & $3.95$ & $34.97$ & $51.22$ & $90.56$ & $74.80$ & $83.50$ & $84.56$ & $59.88$ & $91.50$ \\
  & \textbf{ScaleFL~\cite{scalefl}} & $3.88$ & $34.80$ & $50.95$ & $90.75$ & $75.97$ & $83.28$ & $84.90$ & $60.00$ & $\underline{91.85}$ \\
 & \textbf{LoRA~\cite{lora}} & $\boldsymbol{2.00}$ & $\underline{0.56}$ & $50.05$ & $90.57$ & $74.96$ & $83.59$ & $83.90$ & $58.19$ & $91.13$ \\
  & \textbf{HetLoRA~\cite{hetlora}} & $2.09$ & $0.60$ & $\underline{53.78}$ & $\underline{92.40}$ & $75.16$ & $84.70$ & $85.78$ & $60.04$ & $91.04$ \\
 \rowcolor{gray!30}
 & \textbf{FedHFT} & $\boldsymbol{2.00}$ & $\boldsymbol{0.44}$ & $53.76$ & $\boldsymbol{92.66}$ & $\underline{77.94}$ & $\boldsymbol{84.98}$ & $\boldsymbol{86.35}$ & $\boldsymbol{62.90}$ & $\boldsymbol{91.92}$ \\ \bottomrule
\end{tabular}
\end{adjustbox}
\vspace{0.2cm}
\caption{Performance (matthew's correlation for CoLA and accuracy for the rest) and cost for various methods on various datasets with BERT-base (110M) and BERT-large (340M), for $K=50$, $\alpha=5$.
Columns 3-4 compare the average memory footprint during finetuning and total download+upload size per client respectively.
}
\label{tab:main}
\end{table*}

\begin{table}[]
\begin{adjustbox}{width=\columnwidth}
\begin{tabular}{clcccc}
\toprule
\multirow{2}{*}{Model} & \multirow{2}{*}{\begin{tabular}[c]{@{}l@{}}Finetuning \\ Method\end{tabular}} & \multicolumn{2}{c}{Cost (GB) ($\downarrow$)} & \multicolumn{2}{c}{Score on SQuAD ($\uparrow$)} \\
 &  & Memory & Comm. & EM & F1 \\ \midrule
\multirow{7}{*}{T5-base} & \textbf{FedAvg~\cite{fedavg}} & $4.27$ & $35.60$ & $80.14$ & $85.63$ \\
 & \textbf{FedProx~\cite{fedprox}} & $4.27$ & $35.60$ & $\boldsymbol{82.00}$ & $\boldsymbol{88.17}$ \\
 & \textbf{SCAFFOLD~\cite{scaffold}} & $4.27$ & $71.20$ & $81.65$ & $87.90$ \\
 & \textbf{FedCAM~\cite{fedcam}} & $4.27$ & $35.60$ & $81.93$ & $87.78$ \\
 \cdashline{2-6}\noalign{\vskip 0.4ex}
 & \textbf{DGC~\cite{dgc}} & $4.27$ & $17.97$ & $80.08$ & $85.35$ \\
 & \textbf{HeteroFL~\cite{heterofl}} & $2.67$ & $11.23$ & $79.78$ & $85.11$ \\
 & \textbf{LoRA~\cite{lora}} & $\boldsymbol{1.35}$ & $\underline{0.38}$ & $80.20$ & $85.56$ \\
  & \textbf{HetLoRA~\cite{hetlora}} & ${1.41}$ & ${0.41}$ & $\underline{81.95}$ & $86.96$ \\
 \rowcolor{gray!30}
 & \textbf{FedHFT} & $\boldsymbol{1.35}$ & $\boldsymbol{0.26}$ & $81.88$ & $\underline{87.98}$ \\ \bottomrule
\end{tabular}
\end{adjustbox}
\caption{Performance and cost for various finetuning methods on SQuADv1-dev with T5-base (220M).
}
\vspace{-0.5cm}
\label{tab:t5_base}
\end{table}

\begin{table}[]
\begin{adjustbox}{width=\columnwidth}
\begin{tabular}{clcccc}
\toprule
\multirow{2}{*}{Model} & \multirow{2}{*}{\begin{tabular}[c]{@{}l@{}}Finetuning \\ Method\end{tabular}} & \multicolumn{2}{c}{Cost (GB) ($\downarrow$)} & \multicolumn{2}{c}{Score on SQuAD ($\uparrow$)} \\
 &  & Memory & Comm. & EM & F1 \\ \midrule
\multirow{7}{*}{T5-large} & \textbf{FedAvg~\cite{fedavg}} & $15.10$ & $124.60$ & $82.09$ & $88.86$ \\
 & \textbf{FedProx~\cite{fedprox}} & $15.10$ & $124.60$ & $84.45$ & $91.13$ \\
 & \textbf{SCAFFOLD~\cite{scaffold}} & $15.10$ & $249.20$ & $84.20$ & $91.05$ \\
 & \textbf{FedCAM~\cite{fedcam}} & $15.10$ & $124.60$ & $84.59$ & $\underline{91.14}$ \\
 \cdashline{2-6}\noalign{\vskip 0.4ex}
 & \textbf{DGC~\cite{dgc}} & $15.10$ & $63.15$ & $81.70$ & $88.77$ \\
  & \textbf{HeteroFL~\cite{heterofl}} & $9.50$ & $77.95$ & $81.55$ & $86.13$ \\
 & \textbf{LoRA~\cite{lora}} & $\boldsymbol{5.13}$ & $\underline{1.35}$ & $82.03$ & $88.90$ \\
  & \textbf{HetLoRA~\cite{hetlora}} & ${5.18}$ & ${1.39}$ & $\underline{85.03}$ & $90.95$ \\
 \rowcolor{gray!30}
 & \textbf{FedHFT} & $\boldsymbol{5.13}$ & $\boldsymbol{1.07}$ & $\boldsymbol{85.11}$ & $\boldsymbol{91.33}$ \\ \bottomrule
\end{tabular}
\end{adjustbox}
\caption{Performance and cost for various finetuning methods on SQuADv1-dev with T5-large (770M).
}
\vspace{-0.5cm}
\label{tab:t5_large}
\end{table}

\subsection{Performance/Memory/Communication Efficiency Analysis}
In Table~\ref{tab:main}, we report the results with BERT-base/large for the setting with $K=50$ clients and data heterogeneity ($\alpha=5$). We also report the (i) memory cost - GPU memory footprint due to model weights, gradients, optimizer states, and activations, and (ii) communication cost - total download+upload size for weight updates. For BERT-base/large, we observe that FedHFT obtains the highest score in six of eight datasets with up to 2.46\%/0.80\% higher score compared to the closest performant while achieving 3.09x/3.11x memory cost reduction (memory footprint for finetuning) and 122.57x/126.59x communication cost (size of downloaded and uploaded parameters) reduction compared to FedAvg, FedProx, and FedCAM. We observe consistent results with T5-base/large on SQuADv1 as well, which we report in Tables~\ref{tab:t5_base} and \ref{tab:t5_large}. We observe competitive performance in terms of exact-match (EM) and F1 scores while having significantly lower memory and communication cost compared to the closest performant methods.

\begin{figure}[t!]
    \centering
    \includegraphics[width=\columnwidth]{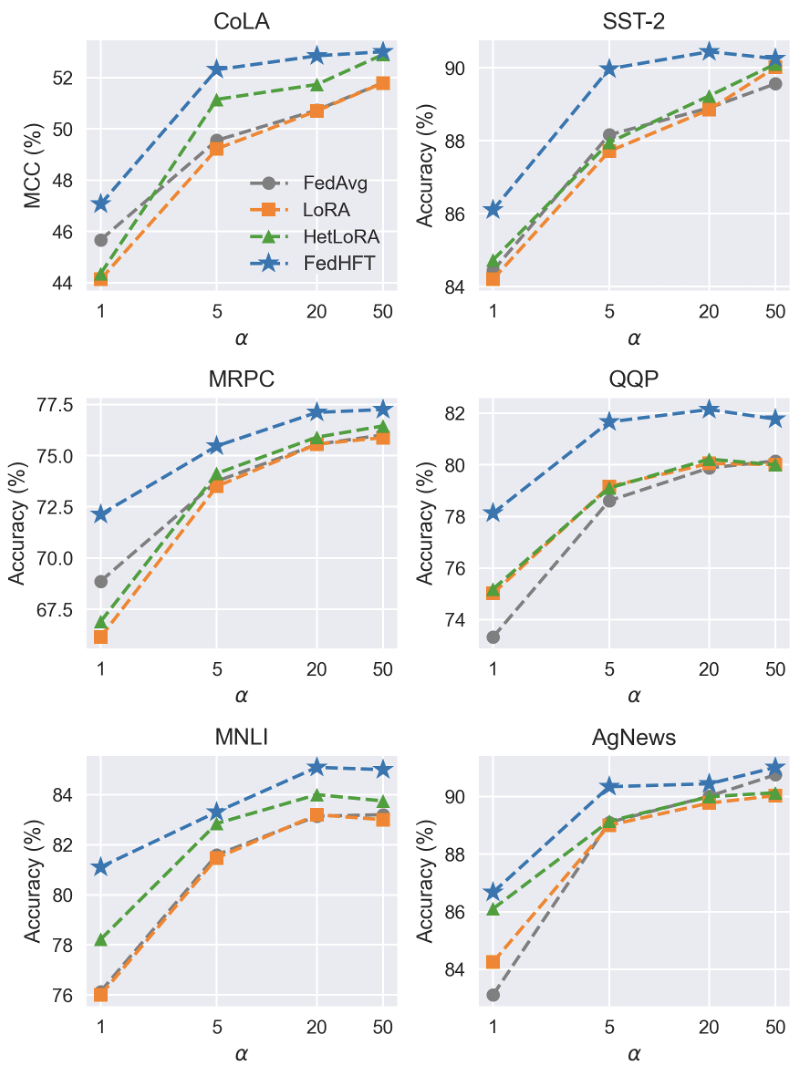}
    \vspace{-20pt}
    \caption{Performance of various resource-efficient finetuning methods and FedAVG as baseline under data heterogeneity with BERT-base, $K=50$, and $\alpha \in [1, 5, 20, 50]$. Lower values of $\alpha$ indicate higher data heterogeneity among clients.}
    \label{fig:data_het}
    \vspace{-10pt}
\end{figure}

FedProx, SCAFFOLD, and FedCAM outperform the baseline FedAvg thanks to handling data heterogeneity better. However, these techniques still don't consider resource heterogeneity since they consume high memory during finetuning and require communicating full model weights at each round. For resource heterogeneity, deep gradient compression (DGC) enables reducing the communication cost during upload however, the downloaded weight size and memory finetuning cost are still the same. In HeteroFL and ScaleFL, we consider four resource levels where clients are uniformly distributed and at lower levels have stricter constraints, resulting in memory and communication cost reduction with slight performance loss. Using LoRA reduces both memory and communication significantly. However, both these methods slightly underperform FedAvg in most datasets. Thanks to the mixture of masked adapters, FedHFT allows each client to communicate fewer parameters and enables personalization with masking and client clustering, which helps us handle data and resource heterogeneity for better performance.

\subsection{Performance Analysis under Various Heterogeneity Settings}
\subsubsection{Data Heterogeneity} 
In Figure~\ref{fig:data_het}, we evaluate FedHFT under various data heterogeneity levels for BERT-base. Lower $\alpha$ indicates higher heterogeneity in terms of label distribution whereas higher $\alpha$ results in distributions closer to iid. We observe that FedHFT brings superior performance across all datasets on various data heterogeneity scenarios. As expected, the performance gap narrows as the distribution becomes more homogeneous ($\alpha=50$), and gets significant when heterogeneity is more severe ($\alpha=1$).

\subsubsection{Resource Heterogeneity}

In Figure~\ref{fig:res_het}, we evaluate FedHFT under various resource heterogeneity levels with BERT-base. To simulate the resource heterogeneity, we assign lower adapter ranks to clients with more constraints on memory and communication resources: (i) We divide clients into four groups where each group operates with $r=4, 8, 16$ and $32$, (ii) two groups where each group operates with $r = 16$ and $32$, and (iii) all clients with adapter rank $r=32$. We observe a performance drop (up to 3\%) as the ratio of clients with lower rank increases due to the loss of resolution in updates, especially for clients with $r=4$. However, the clients with higher ranks preserve performance better and FedHFT outperforms other techniques in most settings.

\begin{figure}[t]
    \centering
    \includegraphics[width=\columnwidth]{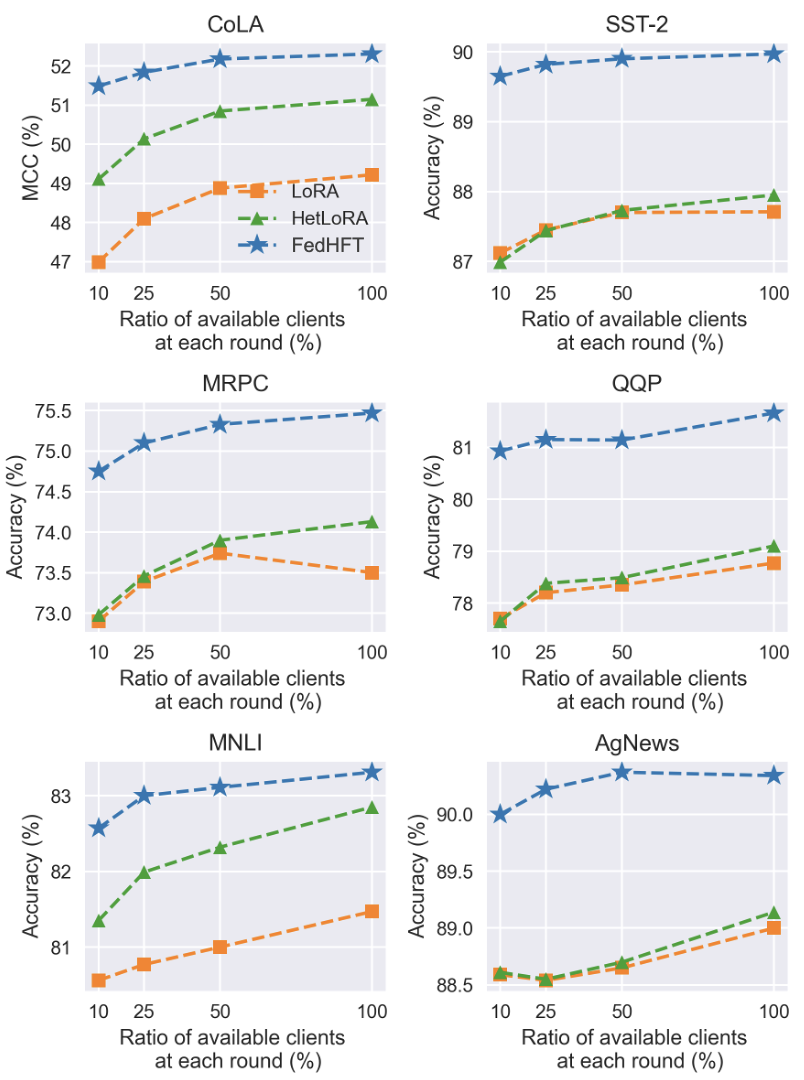}
    \caption{Performance of various federated finetuning methods with BERT-base, $K=50$, $\alpha=5$ under varying client availability ratios $r_a \in [10\%, 25\%, 50\%, 100\%]$.}
    \label{fig:aval_het}
\end{figure}

\subsubsection{Varying Client Availability}
Considering the potential availability, power and connectivity issues of edge clients in real-world applications, we report the results obtained with various client availability ratios. In Figure~\ref{fig:aval_het}, we evaluate FedHFT under various client availability ratios with BERT-base. To simulate this setting, we perform finetuning at each round only on the randomly selected $r_a \%$ of clients. We observe a slight performance drop across all compared methods (up to $\sim 2.4\%$ for FedHFT) as the ratio of unavailable clients increases as expected due to slower and less stable convergence~\cite{fedavg}.

\subsection{Latency Breakdown}
\begin{figure}[t]
    \centering
\includegraphics[width=\columnwidth]{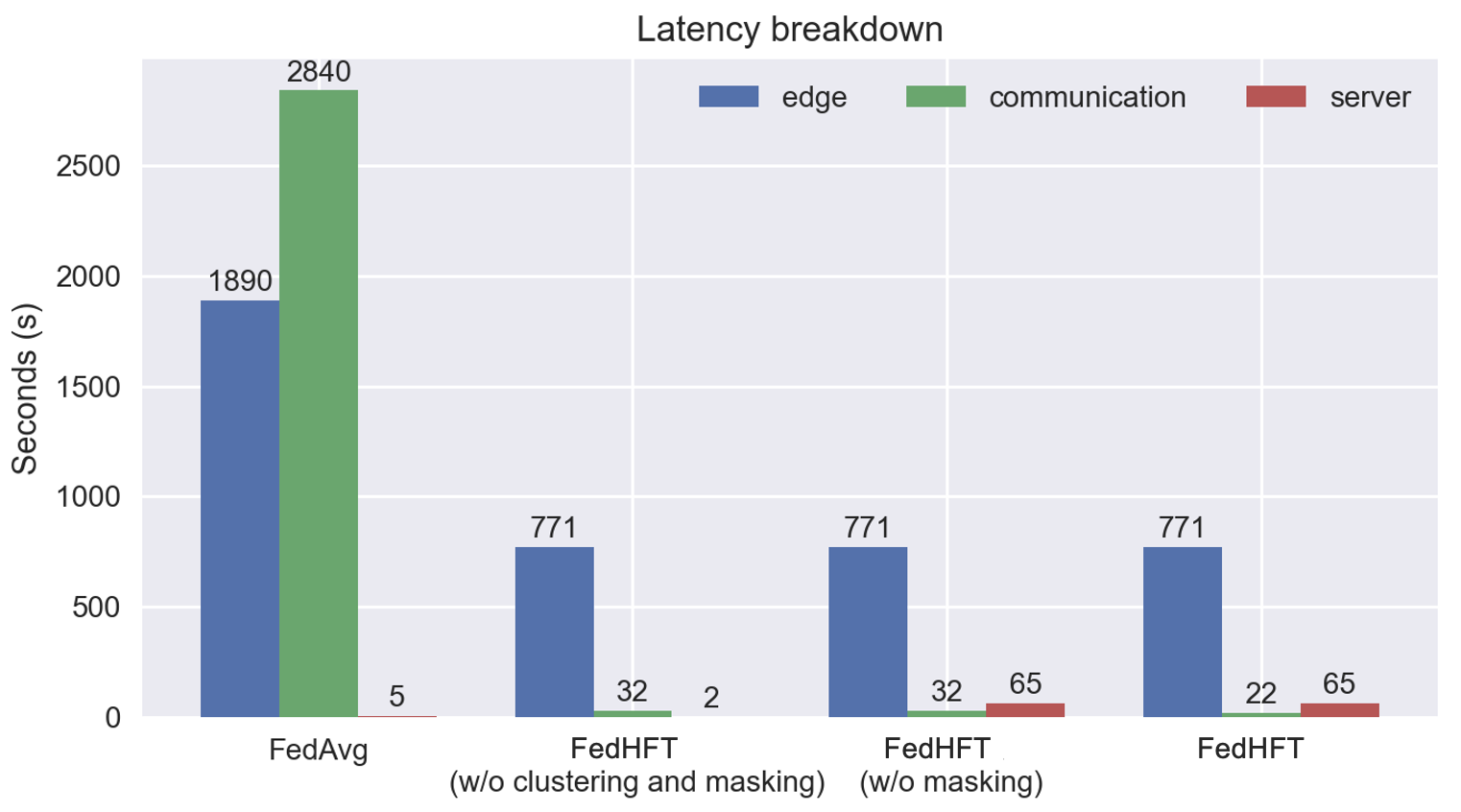}
 \vspace{-20pt}
    \caption{Breakdown of system latency per round (finetuning at edge clients, downloading+uploading parameters, aggregation and clustering at central server) for FedAvg and different variations of FedHFT with BERT-base on SST-2.}
    \label{fig:latency}
     \vspace{-10pt}
\end{figure}

We provide the breakdown of system latency in Figure~\ref{fig:latency} for BERT-base on SST-2. Here, we report the edge (finetuning), communication (upload+download) and server (aggregation and clustering) latency per round obtained with FedAvg and different versions of FedHFT. We note that SVD and clustering take place at the central server, which has extensive resources compared to edge clients. We measure the latency of the aggregation step in BERT-base/large experiments as around 65.2/210.3 seconds on average per round under our implementation (which can be further improved with parallelization techniques), and the latency of communication/local finetuning is inherently much lower (e.g. for BERT-base, $>$100x faster communication, $\sim$2.4x faster finetuning) compared to the standard FL techniques such as FedAvg thanks to adapters and masking.

\begin{table}[]
\begin{adjustbox}{width=\columnwidth}
\begin{tabular}{ccccccc}
\toprule
 \footnotesize{Adapter} & \footnotesize{Cluster} & \footnotesize{Mask} & \footnotesize{CoLA} & \footnotesize{SST-2} & \footnotesize{MRPC} & \footnotesize{AgNews}\\ \midrule
\xmark & \xmark & \xmark & $49.55$ & $88.16$ & $73.75$ & $89.10$ \\
 \cmark & \xmark & \xmark & $49.22$ & $87.71$ & $73.50$ & $89.00$ \\
 \xmark & \cmark & \xmark & $50.97$ & $88.40$ & $74.99$ & $89.76$ \\
 \xmark & \xmark & \cmark & $51.09$ & $88.75$ & $74.65$ & $89.47$ \\
 \cmark & \cmark & \xmark & $51.15$ & $88.34$ & $75.00$ & $89.81$ \\
 \cmark & \xmark & \cmark & $50.89$ & $88.70$ & $74.60$ & $89.45$ \\
 \xmark & \cmark & \cmark & $52.34$ & $90.03$ & $75.51$ & $90.32$ \\
 \cmark & \cmark & \cmark & $52.31$ & $89.97$ & $75.47$ & $90.34$ \\ \bottomrule
\end{tabular}
\end{adjustbox}
\caption{Ablation results of FedHFT with BERT-base to analyze the impact of adapters, clustering, and masking.}
\vspace{-0.5cm}
\label{tab:ablation}
\end{table}
\begin{table}[t]
\begin{adjustbox}{width=\columnwidth}
\begin{tabular}{ccccccc}
\toprule
 Number of Clusters ($C$) & \footnotesize{CoLA} & \footnotesize{SST-2} & \footnotesize{MRPC} & \footnotesize{AgNews}\\ \midrule
 $1$ & $50.89$ & $88.70$ & $74.60$ & $89.45$ \\
 $2$ & $51.77$ & $88.44$ & $75.45$ & $90.15$ \\
 $3$ & $52.31$ & $89.97$ & $75.47$ & $90.34$ \\ 
 $5$ & $52.44$ & $89.90$ & $75.35$ & $90.37$ \\
 $10$ & $52.33$ & $89.62$ & $75.50$ & $90.28$ \\ \bottomrule
\end{tabular}
\end{adjustbox}
\caption{FedHFT with BERT-base using different number of clusters.}
\vspace{0cm}
\label{tab:cluster}
\end{table}

\begin{figure}[t]
    \centering
    \includegraphics[width=\columnwidth]{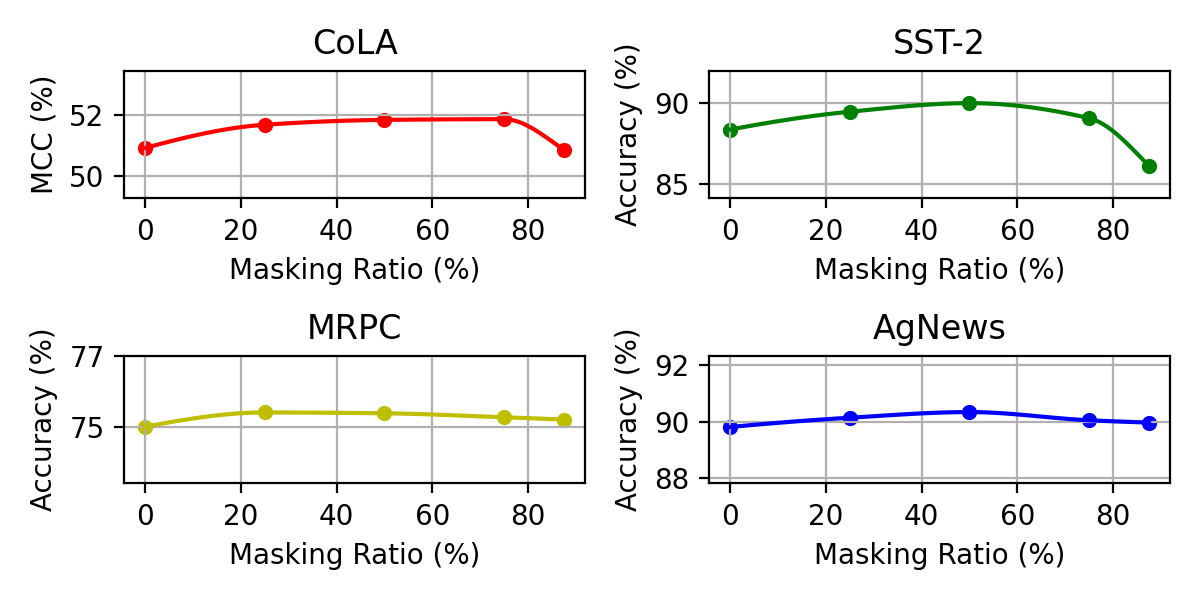}
    \vspace{-0.5cm}
    \caption{Performance of FedHFT with BERT-base for various masking ratios.}
    \label{fig:mask}
\end{figure}

\begin{figure}[t]
    \centering
    \includegraphics[width=\columnwidth]{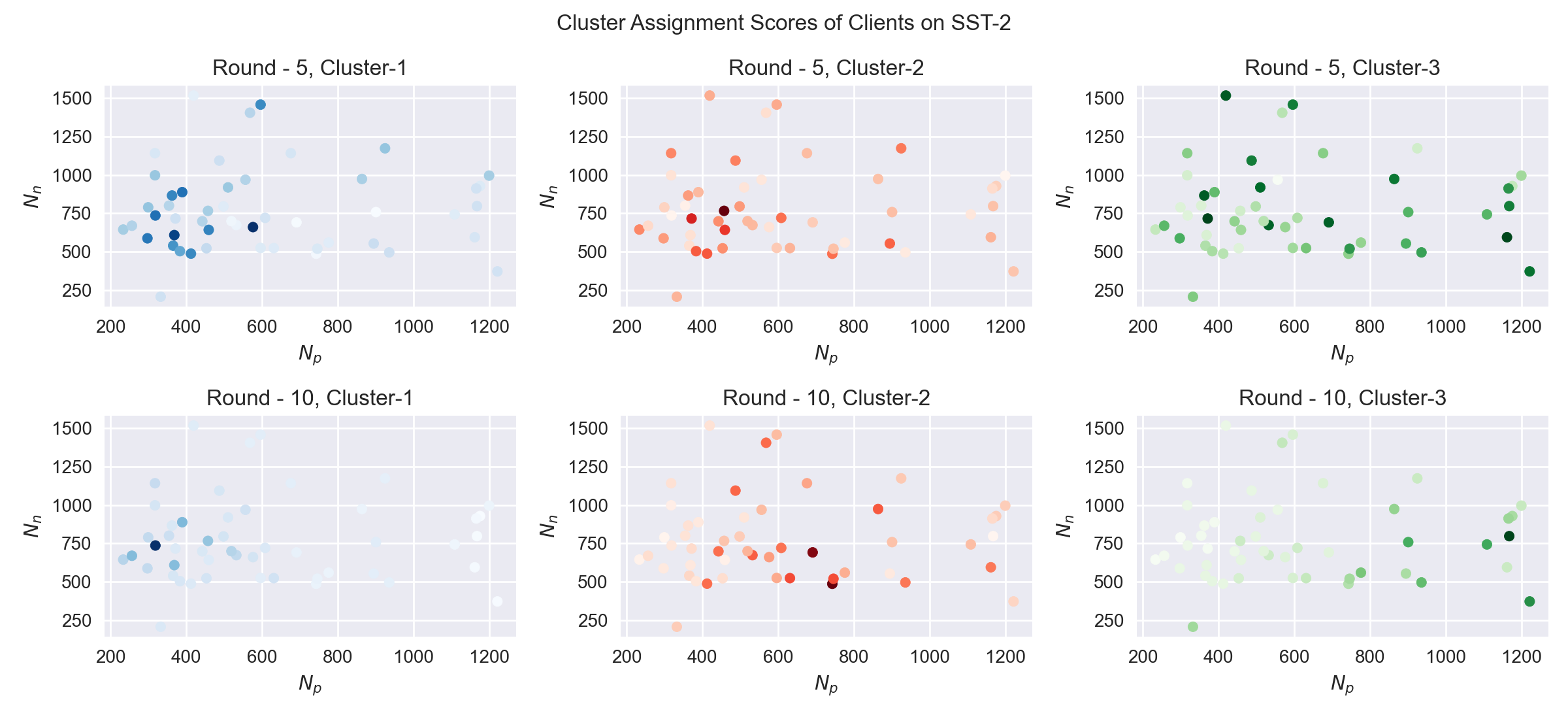}
    \vspace{-0.5cm}
    \caption{Cluster assignment scores with respect to label distribution at rounds 5 and 10 with FedHFT on SST-2. Darker color tones represent higher scores for the respective cluster assignment. x and y axes represent the number of data samples from positive/negative classes in the client's training dataset.}
    \label{fig:cluster}
\end{figure}

 \vspace{-4pt}
\subsection{Ablation Studies}
 \vspace{-2pt}
In Table~\ref{tab:ablation}, we investigate the impact of each component (adapters, clustering and masking) in FedHFT. We observe that clustering and masking improve the performance while using adapters slightly hurts the final model quality. Utilizing the mixture of adapters enables us to combine the performance gains of clustering with the cost efficiency that comes with adapters. Lastly, combining personalized masking with mixture of adapters further enhances performance consistently. 

\begin{table}[b!]
\centering
\def\arraystretch{1}
\begin{adjustbox}{width=\columnwidth}
\begin{tabular}{|c|p{3cm}|p{2.2cm}|p{1.75cm}|p{1.75cm}|}
\hline
\textbf{Dataset} & \textbf{Sentence} & \textbf{Label Distribution at the Client} & \textbf{FT with LoRA} & \textbf{FedHFT} \\ \hline
SST2 & With its dogged hollywood naturalism and the inexorable passage of its characters toward sainthood, windtalkers is nothing but a sticky-sweet soap. (negative) & positive:77\% \quad\quad negative:23\% & positive & negative \\ \hline
SST2 & Writer/director joe carnahan's grimy crime drama is a manual of precinct cliches, but it moves fast enough to cover its clunky dialogue and lapses in logic. (positive) & positive:44\% \quad\quad negative:56\% & negative & positive \\ \hline
CoLA & Mary was told to bring something to the party, so she asked Sue what to. (unacceptable) & acceptable:61\% \quad\quad\quad\quad\quad unacceptable:39\% & acceptable & unacceptable \\ \hline
CoLA & When Alison and David soaked their feet was after dinner. (acceptable) & acceptable:27\% \quad\quad\quad\quad unacceptable:73\% & unacceptable & acceptable \\ \hline
MNLI & Sentence 1: The results of the sheepshead minnow, Cyprinodon variegatus, inland silverside, Menidia beryllina, or mysid, Mysidopsis bahia, tests are acceptable if survival in the controls is 80 percent or greater. Sentence 2: Tests are only acceptable when survival rates during controls come to at least 70 percent. (contradiction) & entailment:22\% \quad\; neutral:55\% \quad\quad contradiction:23\% & neutral & contradiction \\ \hline
MNLI & Sentence 1: Generally, FGD systems tend to be constructed closer to the ground compared to SCR technology retrofits. Sentence 2: SCR technology retrofits differ from FGD systems. (entailment) & entailment:38\% \quad\; neutral:40\% \quad\quad contradiction:22\% & neutral & entailment \\ \hline
\end{tabular}
\label{tab:examples}
\end{adjustbox}
\caption{Example sentences with labels and predictions obtained with BERT-base using FT w/ LoRA and FedHFT.}
\end{table}

We also analyze the impact of the masking ratio. Figure~\ref{fig:mask} reports the results with $r \in [0, 0.875]$. We observe that the overall performance is robust to the selection of this parameter in the range of $[0.25, 0.75]$. Next, Figure~\ref{fig:cluster} illustrates the evolution of cluster assignment scores, showing that the clusters gradually confine. Finally, we analyze the impact of the number of clusters on the performance of FedHFT in Table~\ref{tab:cluster} and observe robustness with no significant performance change after $C>3$ on the analyzed datasets.

\subsection{Example Cases}

In Table~\ref{tab:examples}, we analyze examples with BERT-base ($K=50$, $\alpha=5$) where FedHFT provides correct prediction while federated finetuning with LoRA makes mistakes. These examples illustrate how clustering and masking enable preserving model output quality under data heterogeneity in certain cases. In particular, we observe that using pure LoRA without any technique in FedHFT causes mistakes in difficult samples and the model at the client tends to predict based on the majority data label/distribution in its local training data, and fails under data heterogeneity. Below, we also provide six examples from SQUAD obtained with T5-base using FT with LoRA and FedHFT. Overall, we observe slightly more detailed responses with FedHFT.

\begin{tcolorbox}[breakable,toprule at break=0mm,bottomrule at break=0mm, boxsep=1pt,left=2pt,right=2pt,top=2pt,bottom=2pt]
\scriptsize
\textbf{Text: } OPEC soon lost its preeminent position, and in 1981, its production was surpassed by that of other countries. Additionally, its own member nations were divided. Saudi Arabia, trying to recover market share, increased production, pushing prices down, shrinking or eliminating profits for high-cost producers. The world price, which had peaked during the 1979 energy crisis at nearly \$40 per barrel, decreased during the 1980s to less than \$10 per barrel. Adjusted for inflation, oil briefly fell back to pre-1973 levels. This "sale" price was a windfall for oil-importing nations, both developing and developed.	

\textbf{Question: }  \textit{Which year did the price of oil drop to \$10 per barrel?	
}
\par
\textbf{Answer 1 (FT w/ LoRA):} 1980s

\par
\textbf{Answer 2 (FedHFT):} during the 1980s

\par
\textbf{Question: }  \textit{Why did Saudi Arabia try to increase production, and reduce profits for high cost producers?}
\par
\textbf{Answer 1 (FT w/ LoRA):} market share

\par
\textbf{Answer 2 (FedHFT):} recover market share

\par
\end{tcolorbox}
\begin{tcolorbox}[breakable,toprule at break=0mm,bottomrule at break=0mm, boxsep=1pt,left=2pt,right=2pt,top=2pt,bottom=2pt]
\scriptsize
\textbf{Text: } Effects of inequality researchers have found include higher rates of health and social problems, and lower rates of social goods, a lower level of economic utility in society from resources devoted on high-end consumption, and even a lower level of economic growth when human capital is neglected for high-end consumption. For the top 21 industrialised countries, counting each person equally, life expectancy is lower in more unequal countries (r = -.907). A similar relationship exists among US states (r = -.620).	

\textbf{Question: }  \textit{Why does a lower level of economic growth occur due to high-end consumption?}
\par
\textbf{Answer 1 (FT w/ LoRA):} economic utility

\par
\textbf{Answer 2 (FedHFT):} lower level of economic utility

\textbf{Question: }  \textit{What is a lower rate of social goods an effect of?	
}
\par
\textbf{Answer 1 (FT w/ LoRA):} inequality

\par
\textbf{Answer 2 (FedHFT):} inequality

\par
\end{tcolorbox}

\begin{tcolorbox}[breakable,toprule at break=0mm,bottomrule at break=0mm, boxsep=1pt,left=2pt,right=2pt,top=2pt,bottom=2pt]
\scriptsize
\textbf{Text: } Governor Vaudreuil, who harboured ambitions to become the French commander in chief (in addition to his role as governor), acted during the winter of 1756 before those reinforcements arrived. Scouts had reported the weakness of the British supply chain, so he ordered an attack against the forts Shirley had erected at the Oneida Carry. In the March Battle of Fort Bull, French forces destroyed the fort and large quantities of supplies, including 45,000 pounds of gunpowder. They set back any British hopes for campaigns on Lake Ontario, and endangered the Oswego garrison, already short on supplies. French forces in the Ohio valley also continued to intrigue with Indians throughout the area, encouraging them to raid frontier settlements. This led to ongoing alarms along the western frontiers, with streams of refugees returning east to get away from the action.	

\par
\textbf{Question: }  \textit{Where was there a weakness in British supply chain?	
}
\par
\textbf{Answer 1 (FT w/ LoRA):} supply chain

\par
\textbf{Answer 2 (FedHFT):} Oneida Carry

\textbf{Question: }  \textit{What plans of the British did this attach on Oneida Carry set back?}
\par
\textbf{Answer 1 (FT w/ LoRA):} hopes for campaigns on Lake Ontario

\par
\textbf{Answer 2 (FedHFT):} hopes for campaigns on Lake Ontario, and endangered the Oswego garrison

\par
\end{tcolorbox}

\section{Conclusion}
We have introduced FedHFT, a novel and efficient federated finetuning framework for large language models. Our method utilizes a mixture of masked adapters to reduce memory and communication costs while providing superior performance under data and resource heterogeneity compared to representative techniques. First, adapters combined with masking enable up to $\sim$136x communication and $\sim$3x memory cost reduction, which enables incorporating resource-constrained edge clients into the process. Second, utilizing a mixture of adapters with masking enhances personalization and provides superior performance under heterogeneous data. We have reported the results obtained on extensive experiments and also analyzed the behavior of FedHFT under various heterogeneity settings. Lastly, FedHFT provides a memory and communication cost-efficient solution for edge clients participating in the federated finetuning process but, it introduces workloads on the central server for the storage and update of adapter weights corresponding to each cluster. At the end of each round, the cluster membership scores of clients are updated based on weight updates. Due to the high resource availability and elasticity in cloud data centers where servers are usually hosted in most real-world applications, these additional processes would not cause bottleneck, but are still worth consideration.

\section*{Acknowledgment}
This research is partially sponsored by the NSF CISE grants 2302720, 2312758, 2038029, NSF grant SBE/HNDS 2024320, an IBM faculty award, a grant from CISCO Edge AI program, and Georgia Tech Foundation through the John P. Imlay, Jr. Chair endowment.

\bibliographystyle{abbrv}
\bibliography{refs}

\begin{thebibliography}{10}

\bibitem{gpt3}
T.~Brown, B.~Mann, N.~Ryder, M.~Subbiah, J.~D. Kaplan, P.~Dhariwal, A.~Neelakantan, P.~Shyam, G.~Sastry, A.~Askell, S.~Agarwal, A.~Herbert-Voss, G.~Krueger, T.~Henighan, R.~Child, A.~Ramesh, D.~Ziegler, J.~Wu, C.~Winter, C.~Hesse, M.~Chen, E.~Sigler, M.~Litwin, S.~Gray, B.~Chess, J.~Clark, C.~Berner, S.~McCandlish, A.~Radford, I.~Sutskever, and D.~Amodei.
\newblock Language models are few-shot learners.
\newblock In H.~Larochelle, M.~Ranzato, R.~Hadsell, M.~Balcan, and H.~Lin, editors, {\em Advances in Neural Information Processing Systems}, volume~33, pages 1877--1901, 2020.

\bibitem{fedprompt}
T.~Che, J.~Liu, Y.~Zhou, J.~Ren, J.~Zhou, V.~Sheng, H.~Dai, and D.~Dou.
\newblock Federated learning of large language models with parameter-efficient prompt tuning and adaptive optimization.
\newblock In H.~Bouamor, J.~Pino, and K.~Bali, editors, {\em Proceedings of the 2023 Conference on Empirical Methods in Natural Language Processing}, pages 7871--7888, Singapore, Dec. 2023. Association for Computational Linguistics.

\bibitem{hetlora}
Y.~J. Cho, L.~Liu, Z.~Xu, A.~Fahrezi, and G.~Joshi.
\newblock Heterogeneous lora for federated fine-tuning of on-device foundation models, 2024.

\bibitem{bert}
J.~Devlin, M.-W. Chang, K.~Lee, and K.~Toutanova.
\newblock {BERT}: Pre-training of deep bidirectional transformers for language understanding.
\newblock In J.~Burstein, C.~Doran, and T.~Solorio, editors, {\em Proceedings of the 2019 Conference of the North {A}merican Chapter of the Association for Computational Linguistics: Human Language Technologies, Volume 1 (Long and Short Papers)}, pages 4171--4186, Minneapolis, Minnesota, June 2019. Association for Computational Linguistics.

\bibitem{heterofl}
E.~Diao, J.~Ding, and V.~Tarokh.
\newblock Hetero{\{}fl{\}}: Computation and communication efficient federated learning for heterogeneous clients.
\newblock In {\em International Conference on Learning Representations}, 2021.

\bibitem{peft}
N.~Ding, Y.~Qin, G.~Yang, F.~Wei, Y.~Zonghan, Y.~Su, S.~Hu, Y.~Chen, C.-M. Chan, W.~Chen, J.~Yi, W.~Zhao, X.~Wang, Z.~Liu, H.-T. Zheng, J.~Chen, Y.~Liu, J.~Tang, J.~Li, and M.~Sun.
\newblock Parameter-efficient fine-tuning of large-scale pre-trained language models.
\newblock {\em Nature Machine Intelligence}, 5:1--16, 03 2023.

\bibitem{mrpc}
W.~B. Dolan and C.~Brockett.
\newblock Automatically constructing a corpus of sentential paraphrases.
\newblock In {\em Proceedings of the Third International Workshop on Paraphrasing ({IWP}2005)}, 2005.

\bibitem{ifca}
A.~Ghosh, J.~Chung, D.~Yin, and K.~Ramchandran.
\newblock An efficient framework for clustered federated learning.
\newblock {\em IEEE Transactions on Information Theory}, 68(12):8076--8091, 2022.

\bibitem{grad_sparse}
P.~Han, S.~Wang, and K.~K. Leung.
\newblock Adaptive gradient sparsification for efficient federated learning: An online learning approach.
\newblock In {\em 2020 IEEE 40th International Conference on Distributed Computing Systems (ICDCS)}, pages 300--310, 2020.

\bibitem{adapter}
N.~Houlsby, A.~Giurgiu, S.~Jastrzebski, B.~Morrone, Q.~De~Laroussilhe, A.~Gesmundo, M.~Attariyan, and S.~Gelly.
\newblock Parameter-efficient transfer learning for {NLP}.
\newblock In K.~Chaudhuri and R.~Salakhutdinov, editors, {\em Proceedings of the 36th International Conference on Machine Learning}, volume~97 of {\em Proceedings of Machine Learning Research}, pages 2790--2799. PMLR, 09--15 Jun 2019.

\bibitem{lora}
E.~J. Hu, Y.~Shen, P.~Wallis, Z.~Allen-Zhu, Y.~Li, S.~Wang, L.~Wang, and W.~Chen.
\newblock Lo{RA}: Low-rank adaptation of large language models.
\newblock In {\em International Conference on Learning Representations}, 2022.

\bibitem{llmadapters}
Z.~Hu, Y.~Lan, L.~Wang, W.~Xu, E.-P. Lim, R.~K.-W. Lee, L.~Bing, and S.~Poria.
\newblock Llm-adapters: An adapter family for parameter-efficient fine-tuning of large language models.
\newblock {\em arXiv preprint arXiv:2304.01933}, 2023.

\bibitem{scalefl}
F.~Ilhan, G.~Su, and L.~Liu.
\newblock Scalefl: Resource-adaptive federated learning with heterogeneous clients.
\newblock In {\em Proceedings of The IEEE / CVF Computer Vision and Pattern Recognition Conference (CVPR)}, 2023.

\bibitem{recap}
F.~Ilhan, G.~Su, S.~F. Tekin, T.~Huang, S.~Hu, and L.~Liu.
\newblock Resource-efficient transformer pruning for finetuning of large models.
\newblock In {\em Proceedings of the IEEE/CVF Conference on Computer Vision and Pattern Recognition (CVPR)}, pages 16206--16215, June 2024.

\bibitem{scalefl_2}
F.~Ilhan, G.~Su, Q.~Wang, and L.~Liu.
\newblock Scalable federated learning with system heterogeneity.
\newblock In {\em 2023 IEEE 43rd International Conference on Distributed Computing Systems (ICDCS)}, pages 1037--1040, 2023.

\bibitem{mofl}
Y.~Kang, H.~Gu, X.~Tang, Y.~He, Y.~Zhang, J.~He, Y.~Han, L.~Fan, K.~Chen, and Q.~Yang.
\newblock Optimizing privacy, utility, and efficiency in a constrained multi-objective federated learning framework.
\newblock {\em ACM Trans. Intell. Syst. Technol.}, 15(6), Dec. 2024.

\bibitem{scaffold}
S.~P. Karimireddy, S.~Kale, M.~Mohri, S.~Reddi, S.~Stich, and A.~T. Suresh.
\newblock {SCAFFOLD}: Stochastic controlled averaging for federated learning.
\newblock In H.~D. III and A.~Singh, editors, {\em Proceedings of the 37th International Conference on Machine Learning}, volume 119 of {\em Proceedings of Machine Learning Research}, pages 5132--5143. PMLR, 13--18 Jul 2020.

\bibitem{fedprox}
T.~Li, A.~K. Sahu, M.~Zaheer, M.~Sanjabi, A.~Talwalkar, and V.~Smith.
\newblock Federated optimization in heterogeneous networks.
\newblock In I.~S. Dhillon, D.~S. Papailiopoulos, and V.~Sze, editors, {\em Proceedings of Machine Learning and Systems 2020, MLSys 2020, Austin, TX, USA, March 2-4, 2020}. mlsys.org, 2020.

\bibitem{prefix}
X.~L. Li and P.~Liang.
\newblock Prefix-tuning: Optimizing continuous prompts for generation.
\newblock In C.~Zong, F.~Xia, W.~Li, and R.~Navigli, editors, {\em Proceedings of the 59th Annual Meeting of the Association for Computational Linguistics and the 11th International Joint Conference on Natural Language Processing (Volume 1: Long Papers)}, pages 4582--4597, Online, Aug. 2021. Association for Computational Linguistics.

\bibitem{dgc}
Y.~Lin, S.~Han, H.~Mao, Y.~Wang, and W.~J. Dally.
\newblock {Deep Gradient Compression: Reducing the communication bandwidth for distributed training}.
\newblock In {\em The International Conference on Learning Representations}, 2018.

\bibitem{fedcam}
J.~Ma, T.~Zhou, G.~Long, J.~Jiang, and C.~Zhang.
\newblock Structured federated learning through clustered additive modeling.
\newblock In A.~Oh, T.~Naumann, A.~Globerson, K.~Saenko, M.~Hardt, and S.~Levine, editors, {\em Advances in Neural Information Processing Systems}, volume~36, pages 43097--43107. Curran Associates, Inc., 2023.

\bibitem{fedavg}
B.~McMahan, E.~Moore, D.~Ramage, S.~Hampson, and B.~A.~y. Arcas.
\newblock {Communication-Efficient Learning of Deep Networks from Decentralized Data}.
\newblock In A.~Singh and J.~Zhu, editors, {\em Proceedings of the 20th International Conference on Artificial Intelligence and Statistics}, volume~54 of {\em Proceedings of Machine Learning Research}, pages 1273--1282. PMLR, 20--22 Apr 2017.

\bibitem{t5}
C.~Raffel, N.~M. Shazeer, A.~Roberts, K.~Lee, S.~Narang, M.~Matena, Y.~Zhou, W.~Li, and P.~J. Liu.
\newblock Exploring the limits of transfer learning with a unified text-to-text transformer.
\newblock {\em J. Mach. Learn. Res.}, 21:140:1--140:67, 2019.

\bibitem{squad}
P.~Rajpurkar, J.~Zhang, K.~Lopyrev, and P.~Liang.
\newblock {SQ}u{AD}: 100,000+ questions for machine comprehension of text.
\newblock In J.~Su, K.~Duh, and X.~Carreras, editors, {\em Proceedings of the 2016 Conference on Empirical Methods in Natural Language Processing}, pages 2383--2392, Austin, Texas, Nov. 2016. Association for Computational Linguistics.

\bibitem{sst2}
R.~Socher, A.~Perelygin, J.~Wu, J.~Chuang, C.~D. Manning, A.~Ng, and C.~Potts.
\newblock Recursive deep models for semantic compositionality over a sentiment treebank.
\newblock In D.~Yarowsky, T.~Baldwin, A.~Korhonen, K.~Livescu, and S.~Bethard, editors, {\em Proceedings of the 2013 Conference on Empirical Methods in Natural Language Processing}, pages 1631--1642, Seattle, Washington, USA, Oct. 2013. Association for Computational Linguistics.

\bibitem{fl_lora}
Y.~Sun, Z.~Li, Y.~Li, and B.~Ding.
\newblock Improving lo{RA} in privacy-preserving federated learning.
\newblock In {\em The Twelfth International Conference on Learning Representations}, 2024.

\bibitem{sparse_train}
Y.-L. Sung, V.~Nair, and C.~Raffel.
\newblock Training neural networks with fixed sparse masks.
\newblock {\em ArXiv}, abs/2111.09839, 2021.

\bibitem{llama2}
H.~Touvron, L.~Martin, K.~Stone, P.~Albert, A.~Almahairi, Y.~Babaei, N.~Bashlykov, S.~Batra, P.~Bhargava, S.~Bhosale, D.~Bikel, L.~Blecher, C.~C. Ferrer, M.~Chen, G.~Cucurull, D.~Esiobu, J.~Fernandes, J.~Fu, W.~Fu, B.~Fuller, C.~Gao, V.~Goswami, N.~Goyal, A.~Hartshorn, S.~Hosseini, R.~Hou, H.~Inan, M.~Kardas, V.~Kerkez, M.~Khabsa, I.~Kloumann, A.~Korenev, P.~S. Koura, M.-A. Lachaux, T.~Lavril, J.~Lee, D.~Liskovich, Y.~Lu, Y.~Mao, X.~Martinet, T.~Mihaylov, P.~Mishra, I.~Molybog, Y.~Nie, A.~Poulton, J.~Reizenstein, R.~Rungta, K.~Saladi, A.~Schelten, R.~Silva, E.~M. Smith, R.~Subramanian, X.~E. Tan, B.~Tang, R.~Taylor, A.~Williams, J.~X. Kuan, P.~Xu, Z.~Yan, I.~Zarov, Y.~Zhang, A.~Fan, M.~Kambadur, S.~Narang, A.~Rodriguez, R.~Stojnic, S.~Edunov, and T.~Scialom.
\newblock Llama 2: Open foundation and fine-tuned chat models, 2023.

\bibitem{glue}
A.~Wang, A.~Singh, J.~Michael, F.~Hill, O.~Levy, and S.~Bowman.
\newblock {GLUE}: A multi-task benchmark and analysis platform for natural language understanding.
\newblock In T.~Linzen, G.~Chrupa{\l}a, and A.~Alishahi, editors, {\em Proceedings of the 2018 {EMNLP} Workshop {B}lackbox{NLP}: Analyzing and Interpreting Neural Networks for {NLP}}, pages 353--355, Brussels, Belgium, Nov. 2018. Association for Computational Linguistics.

\bibitem{cola}
A.~Warstadt, A.~Singh, and S.~R. Bowman.
\newblock Neural network acceptability judgments.
\newblock {\em Transactions of the Association for Computational Linguistics}, 7:625--641, 2019.

\bibitem{mnli}
A.~Williams, N.~Nangia, and S.~Bowman.
\newblock A broad-coverage challenge corpus for sentence understanding through inference.
\newblock In M.~Walker, H.~Ji, and A.~Stent, editors, {\em Proceedings of the 2018 Conference of the North {A}merican Chapter of the Association for Computational Linguistics: Human Language Technologies, Volume 1 (Long Papers)}, pages 1112--1122, New Orleans, Louisiana, June 2018. Association for Computational Linguistics.

\bibitem{agnews}
X.~Zhang, J.~Zhao, and Y.~LeCun.
\newblock Character-level convolutional networks for text classification.
\newblock In C.~Cortes, N.~Lawrence, D.~Lee, M.~Sugiyama, and R.~Garnett, editors, {\em Advances in Neural Information Processing Systems}, volume~28. Curran Associates, Inc., 2015.

\end{thebibliography}

\end{document}